\documentclass{article}
\usepackage[margin=1.5in]{geometry}

\usepackage{times}
\usepackage{latexsym}
\usepackage[utf8]{inputenc}
\usepackage{graphicx}
\usepackage{tikz}
\usepackage{amsmath}
\usepackage{booktabs}
\usepackage{pgfplots,pgfplotstable}
\usepackage{filecontents}
\pgfplotsset{compat=1.10}

\usepackage{url}

\date{}

\title{A Hybrid Solution to Learn Turn-Taking \\ in Multi-Party Service-based Chat Groups}

\usepackage{authblk}
\author{Ma\'{i}ra Gatti de Bayser} 
\author{Melina Alberio Guerra} 
\author{Paulo Cavalin} 
\author{Claudio Pinhanez}
\affil{IBM Research, Rio de Janeiro, Brazil}

\begin{document}

\label{firstpage}
\maketitle

\begin{abstract}
To predict the next most likely participant to interact in a multi-party conversation is a difficult problem. In a text-based chat group, the only information available is the sender, the content of the text and the dialogue history. In this paper we present our study on how these information can be used on the prediction task through a corpus and architecture that integrates turn-taking classifiers based on Maximum Likelihood Expectation (MLE), Convolutional Neural Networks (CNN) and Finite State Automata (FSA). The corpus is a synthetic adaptation of the Multi-Domain Wizard-of-Oz dataset (MultiWOZ) to a multiple travel service-based bots scenario with dialogue errors and was created to simulate user's interaction and evaluate the architecture. We present experimental results which show that the CNN approach achieves better performance than the baseline with an accuracy of 92.34\%, but the integrated solution with MLE, CNN and FSA achieves performance even better, with 95.65\%.
\end{abstract}

\section{Introduction}
The \textit{multi-party turn-taking problem} consists of determining the proper turn to interact in a conversation with more than two participants. Fundamentally, the goal is \textit{to predict which agent in the conversation is the most likely to speak next} and, conversely, when an agent must wait before interacting. An \textbf{\textit{agent}} can be either a person or a chatbot. That interaction can be a reply to the last interaction, a reply to an interaction in the past in the dialogue, or even an interruption. The two former are replies to existent turns, while the later is the creation of a new turn.

In this paper we present a hybrid architecture that has two components for controlling the dialogue of multiple travel service-based bots with an user. Previous works have implemented turn-taking controlling through a finite-state automata (FSA) based service which is called for every utterance  exchanged in the group chat, by considering both the content and the history of interaction between participants \cite{finch2018}\cite{Emas18}\cite{Deepdial18}. This service is part of a platform called Ravel which enables the connection of several chatbots in a chat group with users. The rule-based system solves the turn-taking problem for an investment advisor scenario but it has limitations for scaling up the set of rules and the application to new domains since it is heavily dependent on expert's knowledge.

Therefore, more recently \cite{GattideBayser19}, machine learning (ML) approaches which may provide a more scalable solution depending on the size of the dataset have also being applied to this problem. Given a finite set of possible agents that can speak, the learned models predict the most likely agent to speak next, assuming only one agent should speak at a time, and the information to predict that can include only participant or both participant and content data. From the tested approaches, although the CNN modeling which considered both content and the agent information achieved the best performance, it requires a lot of data. While the MLE modeling requires less data but still was not as good as CNN. Therefore, there is a need for a hybrid approach.

To achieve that, we have then created a system which is an extension and instantiation of Ravel's architecture. It integrates the MLE, CNN and FSA-based turn-taking classifiers in order to solve the task. To evaluate the system, we have adapted the \textit{multibotwoz} corpus \cite{Budzianowski18}\cite{GattideBayser19} to a multiple travel service-based bots scenario with dialogue errors in order to simulate the user's interaction. We present experimental results which show that the integrated solution with MLE, CNN and FSA achieves performance even better (95.65\% of accuracy) than the solution with only the CNN on the same dataset (92.34\%).

\section{Dataset}

The Multi-Domain Wizard-of-Oz dataset (MultiWOZ)\footnote{MultiWoZ dataset: https://www.repository.cam.ac.uk/han- dle/1810/280608} \cite{Budzianowski18} is a fully-labeled collection of human-human written conversations spanning over multiple domains and topics. This dataset was not created considering that more than one bot would be in the conversation with the user. Rather, it was created considering a dyadic conversation between the user and a bot that can talk about multiple domains or topics, and the topics are actually service providers. 

\begin{figure}[hb!]
\centering
\begin{tikzpicture}[scale=0.85]
\begin{axis}[ybar,legend style={at={(0.5,-0.2)},anchor=north,legend columns=-1},ylabel={Interactions Ratio},symbolic x coords={user, train\_bot, hotel\_bot, attraction\_bot, restaurant\_bot, travel\_bot, taxi\_bot},
xtick=data,nodes near coords,nodes near coords align={vertical},x tick label style={rotate=45,anchor=east},]
\addplot coordinates {
(user, 50)
(train\_bot, 12)
(hotel\_bot, 11)
(restaurant\_bot, 11)
(attraction\_bot, 10)
(travel\_bot, 4)
(taxi\_bot, 2)};
\end{axis}
\end{tikzpicture}
\caption{Bots Interactions Distribution}
\label{fig:bots}
\end{figure}
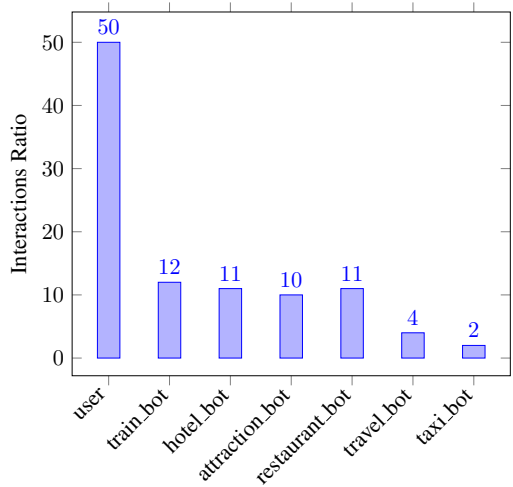

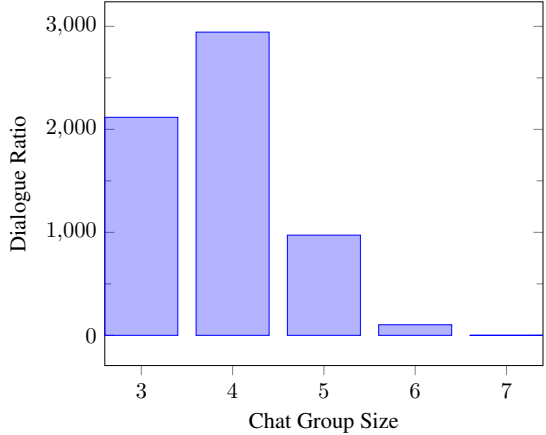
\begin{figure}[ht]
\centering
\begin{tikzpicture}[scale=0.85]
\begin{axis}[
yticklabel style={/pgf/number format/fixed},
scaled y ticks = false,
minor y tick num={1},
xtick pos=left,
legend cell align = left,
legend style={draw=none},
xlabel = {Chat Group Size},
ylabel = {Dialogue Ratio},
ybar
]
\addplot[blue,ybar,fill, fill opacity=0.3, bar width = 0.8] table {data.dat};
\end{axis}
\end{tikzpicture}
\caption{Dialogue Distribution per Chat Group Size}
\label{fig:chatgroup}
\end{figure}

\begin{table}[hbt]
\centering
\small
\begin{tabular}{@{}rrrrcrrr@{}}
\toprule 
 \textbf{Metric}  & \textbf{Value}    
\\
\midrule
\textit{Nbr. of utterances)} & 99,553 \\
\textit{Nbr. of Dialogues} &  6,138 \\ 
\textit{Avg. nbr. of agents per Dialogue}  & 4 (std: 1)  \\
\textit{Avg. nbr. of utterances per Dialogue} &  16 \\
\textit{Avg. length of utterances (words)}  & 13\\
\bottomrule
\end{tabular}
\caption{Multibotwoz Corpus Summary.}
\label{tab:corpus}
\end{table}

An adapted version of this corpus has been presented in \cite{GattideBayser19}. The dialogues with only two topics or that contained topics which are present in small number of dialogues were filtered. The sender was then classified  as one of the topic in order to determine the service provider. I.e., if the utterance was about booking a hotel room, even though there is no mention to the term "hotel" or "room", the sender was labeled as \textit{hotel\_bot}. The resulted corpus was called the \textit{multibotwoz} dataset and it contains only the following services: attractions, hotel, restaurant, taxi, train and travel bot, and ended up with $99,553$ utterances in $6,138$ dialogues with $4$ agents on average in each dialogue, varying from $3$ to $7$ and $16$ utterances exchanged on average in each dialogue (see Table~\ref{tab:corpus}). No bot interacts after another bot in the \textit{multibotwoz} dataset, only after the \textit{user} \footnote{Multibotwoz dataset: https://github.com/CognitiveHorizons/ AIHN-publications}. Figures~\ref{fig:bots} and ~\ref{fig:chatgroup} present more details with the bots interaction ratio compared to the user interaction and the dialogue distribution per chat group size.

\section{Turn-Taking Modeler}

Based on the results on the comparison of ML modeling approaches as presented by \cite{GattideBayser19}, we have chosen the MLE and CNN. We therefore further describe them in this section along with the baseline. 

\subsection{Baseline}

We consider a baseline which we call \textbf{Repeat Last} to compare our proposed methods: this approach is based on a social rule often observed in multi-party human dialogues \cite{Sacks1974}: whenever an agent speaks, we might predict the next one as being the one that had spoken before.  More formally, the \textit{Repeat Last} baseline prediction works as following: let $A=\{a_i  |  1 \leq i \leq n\}$ be the set of agents in the dialogue, $n$ be the number of agents, and let $S = \{s_t  |  1 \leq t \leq T\}$ be the set of agents who sent an utterance in the dialogue up to a time $T$, where $s_t \in A$. 

Whenever the speaker $s_t$ sends an utterance, the next agent selected to talk, denoted $s_{t+1}$, is the one who spoke at time $t-1$, i. e., $s_{t+1} = s_{t-1}$.

\subsection{MLE and CNN Modeling}

We make use of the \textit{one-hot} encoding to convert the information of the agents to a feature vector, formalized as follows. Let $x$ be a vector and $x \in C^n $, a $n$-dimensional instance space with $n$ agents in the conversation and $a_i$ the $i$-th agent, where $1 \leq i \leq n$, and let $s_t$ be the agent who spoke at time $t$. The binary feature vector $x(t)$ at time $t$ of the dialog, can be defined as:

\begin{equation}
\label{eq:1}
 x(t) = 
 \begin{bmatrix} {x_{t}}_1 & {x_{t}}_2 & ... & {x_{t}}_n \end{bmatrix}^T  
 \end{equation}
 \begin{equation}
 {x_t}_i = 
 \begin{cases}
    1  & \quad \text{if } a_i \text{ is the sender, i.e. } a_i = s_t\\
    0  & \quad \text{if } a_i \text{ is not the sender, i.e.} a_i \ne s_t
  \end{cases}
\end{equation}

Therefore, in order to produce the input vector for our models, a linear transformation $T : C^n \rightarrow C^{W * n}$ on $x(t)$ is performed by taking into account $x(t)$ until $x(t-W))$, where $W$ is the size of lookback window  and $t > W$, as:

\begin{equation}
\small
 x'(t) = 
 \begin{bmatrix} {x_{t}}_1 & \ldots & {x_{t}}_n  & {x_{t-W)}}_1 &  \ldots & {x_{t-W)}}_n \end{bmatrix}^T
 \label{eq:binary}
\end{equation}

\noindent \textbf{A-MLE}: The A-MLE applies the \textit{Maximum Likelihood Estimation} \cite{Harris98} after encoding only the information of the agents and by making use of the aforementioned one-hot encoding method. This learning method takes into account only the order in which the agents interact in the conversation. Therefore, transitions are learned by considering that the previous state is the last agent which sent an utterance and the next state is the following agent which sent an utterance. We modeled A-MLE considering a lookback window of size 2, which means the previous state contains information of the two last agents which sent an utterance. In this case, a $\theta$ transition from state $\pi-1$ to $\pi$, is modeled as:
\begin{equation}
\small
\label{eq:bmle}
\theta : state(\pi-1) =  x'(t) \rightarrow state(\pi) = x(t+1)
\end{equation}
We then compute the MLE with smoothing to estimate the parameter for each $\theta (\pi) \in \Theta$ transition type. Therefore, for each corpus, we estimate $L$ for observed transitions as:
\begin{equation}
\small
\label{eq-likelihood}
 L (\theta | x'(t), x(t+1))= 
 \frac{\displaystyle count(\theta,x'(t), x(t+1))+1}{\displaystyle count(\theta,x'(t), x(t+1))+|\Pi|}
\end{equation}
Where $\Pi$ is the set of states and $|\Pi|$ is the number of states in the set.\\

\noindent \textbf{AC-CNN}: the agent-and-content \textit{convolutional neural network (AC-CNN)} modeler consists of a standard CNN modeling used for text classification adapted for the turn-taking task. Such adaptation consists of formatting the previous utterances and the name of the agent as a raw text, and defining the label as in the previous methods. More formally, let $s_{t-1}$ be the agent who spoke utterance $u_{t-1}$ at time $t-1$, and $s_t$ the agent who spoke the last utterance $u_{t}$, to predict who will speak at time $t+1$, we build the following raw text: $s_{t-1} \oplus u_{t-1} \oplus s_t \oplus u_t$, where $\oplus$ represents the concatenation of textual strings. That text is then used as input to the neural network. 

The CNN's architecture was designed with an embedding layer with 64 dimensions; dropout set to 0.2; convolutional layer with 64 filters with kernel size of 3 and stride equals to 1; 1D Global Max-pooling layer with pool size set to 5; another dropout set to 0.2; and 300-dimensional dense hidden layer.

The CNN model does not constraint with regard to waiting for a specific moment to start predicting, it follows a more classical batch-learning process. We considered a $70/30$ train-test split, where $70\%$ of subsequent dialogues are used for training and the remaining $30\%$ for testing. In order to set meta-parameters for the models, cross-validation has been applied on the training set. The vocabulary is built with training and testing data, therefore, all words had WE and there were no words which where OOV. For both the embedding and the hidden layers in the AC-CNN models, \textit{Rectified-Linear-Units activation functions (Relu)} are applied. For the training, we make use of the \textit{Adam optimizer}, with 3 epochs for training and learning rate set to 0.001. Batch size is set to 5.

\begin{table}[htb]
\small
\centering
\begin{tabular}{@{}rrrrcrrr@{}}
\toprule 
 & \textbf{A-MLE} & \textbf{AC-CNN}  
\\
\midrule
Accuracy &  84.39\% & 92.34\% \\ 
Disjoint Errors &  65.20\% & 32.13\% \\ 
\bottomrule
\end{tabular}
\caption{Accuracy and Disjoint Errors.}
\label{tab:accuracyerrors}
\end{table}

Table~\ref{tab:accuracyerrors} presents the accuracy and the percentage of disjoint errors between both A-MLE and AC-CNN modelers. 
Let $E_{A-MLE}$ be the set of errors achieved by A-MLE predictor, and $E_{AC-CNN}$ be the set of errors achieved by AC-CNN. The intersection between the sets ($E_{A-MLE} \cap E_{AC-CNN}$) was only $29.87\%$ of the union of the sets ($E_{A-MLE} \cup E_{AC-CNN}$).
The relative complement of $E_{AC-CNN}$ in $E_{A-MLE}$ ($E_{A-MLE} \setminus E_{AC-CNN}$) was $65.20\%$ of $E_{A-MLE}$, while the relative complement of $E_{A-MLE}$ in $E_{AC-CNN}$ ($E_{AC-CNN} \setminus E_{A-MLE}$) was $32.13\%$ of $E_{AC-CNN}$. Therefore, our proposed solution was defined with both classifiers with the goal to maximize the accuracy.
		
\subsection{FSA-based Turn Taking}

Ravel's platform \cite{Emas18} is a MAS-based micro-services-driven architecture platform that enables the connection of conversational systems in a multi-bot environment. Ravel's environment is mainly composed of an agent which is a Communication Hub (CH) that enables the message exchange between the chatbots which are agents; a Connector, which connects the agents to the CH; and a FSA-based Conversation Governance (CG) service to orchestrate the turn-taking. The CG service is implemented as an interpreter of a Domain Specific Language for Conversation Rules \cite{Deepdial18} (DSL-CR), which enables modeling, specification, and execution of multi-party turn-taking through deontic logic.

The \texttt{Turn} in the conversation is the exchange by one participant (agent or person) of one message which contains one or more utterances. It represents an event which can change the state of the conversation or the set of norms which are active in the conversation such as an utterance arrival. For a given turn in a conversation, the norms are defined  as: 

\noindent \textit{An \textbf{obligation} requires the participant to pro-actively or reactively emit an utterance}; \\
\noindent \textit{A \textbf{permission} allows the participant to pro-actively or reactively emit an utterance};  \\
\noindent \textit{A \textbf{prohibition} forbids the participant to emit utterances, or states that they are not expected in that turn}. 

For each message that arrives, the GC service may use variables as \textit{\$sender}, \textit{\$last\_sender} and \textit{\$receivers}, besides the participant roles, as dialogue context information to identify the members that can be eligible to receive the activated norms: the sender of the message, the sender of the message before the current and the agents which were mentioned in current message (if any), respectively. 

\section{The Proposed Hybrid Solution}

\begin{figure}[ht]
\centerline{\includegraphics[width=8.0cm]{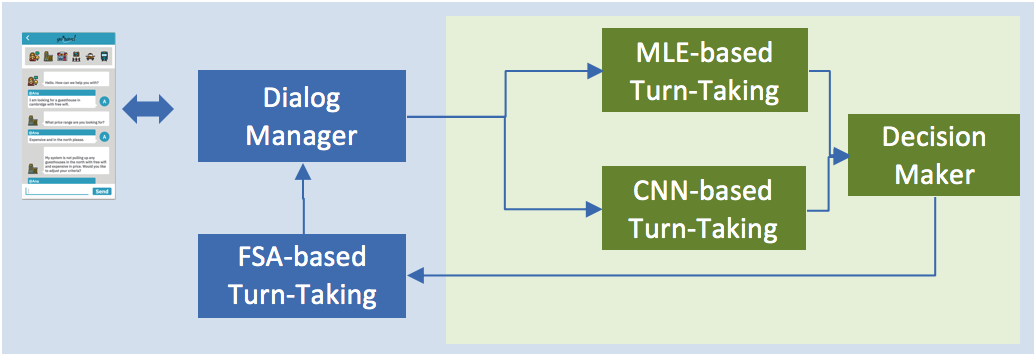}}
\caption{The Proposed Architecture Workflow}
\label{fig:system}
\end{figure}

In our hybrid solution we propose to model the output of the MLE and CNN classifiers into the finite state automata definition which is the input of the FSA-based (GC) service from Ravel, as illustrated in Figure ~\ref{fig:system}.

More formally, the proposed solution prediction works as following: let $A=\{a_i  |  1 \leq i \leq n\}$ be the set of agents in the dialogue, $n$ be the number of agents, and let $S = \{s_t  |  1 \leq t \leq T\}$ be the set of agents who sent an utterance in the dialogue up to a time $T$, where $s_t \in A$. 
  
Whenever the speaker $s_t$ sends an utterance, the next agent selected to talk, denoted $s_{t+1}$, is retrieved as:

 \begin{equation}
 s_{t+1} = \ell(x(t+1)) | \ell \in L
 \end{equation}
 
 \begin{equation}
 \small \small
 \ell(x(t+1)) = 
 \begin{cases}
    \ell(x_{1}(t+1)) & \quad \text{if } C_{1} >= k_{1}\\
    \ell(x_{2}(t+1)) & \quad \text{if } C_{2} < k_{1}\\ 
                            & \quad \textrm{and}  \quad C_{2} >= k_{2}\\
    travel\_bot  & \quad otherwise
  \end{cases}
  \label{eq:prediction}
\end{equation}

Where $C_{1}$ and $C_{2}$ are the confidence scores for prediction using AC-CNN and A-MLE, respectively, and $k_{1}$ and $k_{2}$ are thresholds for each classifier.

The FSA-based Turn-Taking acts as a binary classifier by deciding, for a given input with sender, content (the utterance), i.e., $x'(t)$ and the predicted sender ($\ell(x(t+1))$), if the current sender can or cannot interact in that turn.

\section{Experimental Results}

\begin{figure}[h]
\centerline{\includegraphics[width=8.0cm]{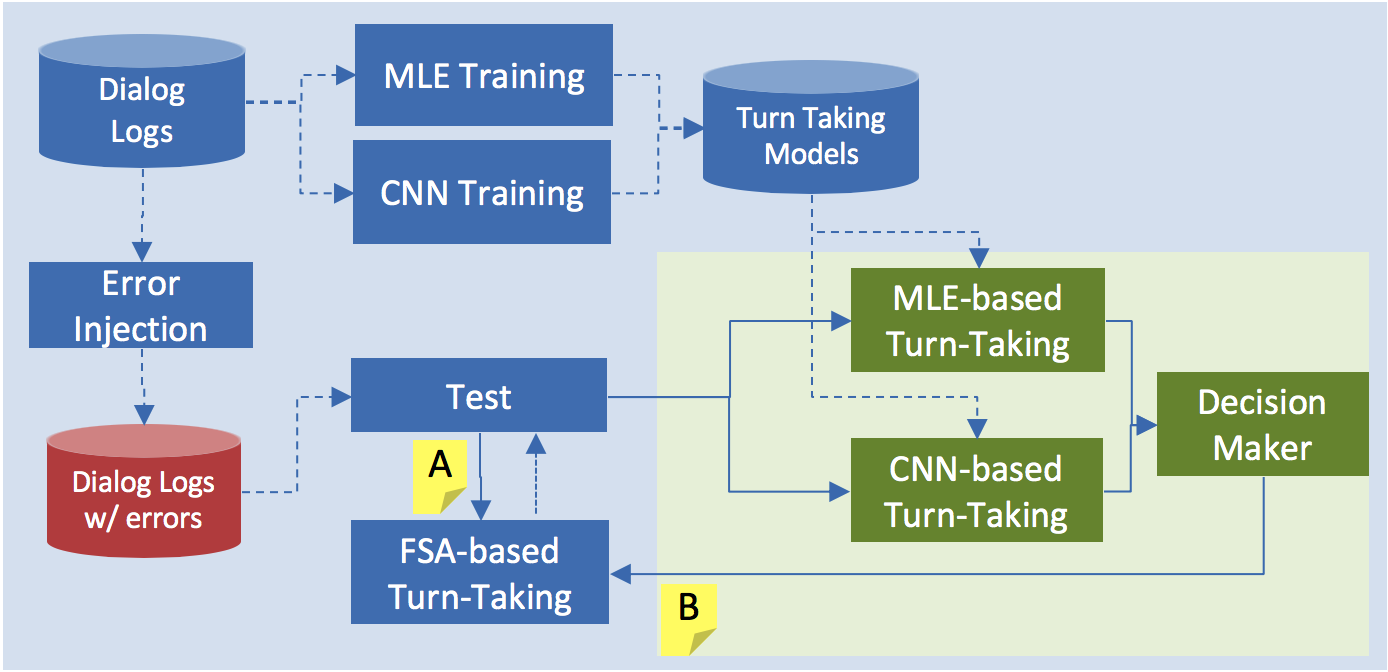}}
\caption{Validation Scenarios}
\label{fig:scenarios}
\end{figure}

In a live chat between humans and chatbots, it is not possible to determine when the bots tries to reply to the user's utterance, hence nor the order. A bot that is not supposed to interact (for instance, because the user is requesting information about a train and not a taxi), should not have its response broadcasted in the group. Because of that, for our tests datasets, we have extended the \textit{multibotwoz} corpus\footnote{Multibotwoz Corpus with Dialogue Errors: https:// github.com/CognitiveHorizons/AIHN-publications} to include dialogue errors in order to simulate the interactions from bots that are not supposed to interact in a given turn and which try to do so (see Table~\ref{tab:corpuserrors}). Therefore, for each correct answer of one bot to the previous sentence sent by the user, we added another answer from the other bots. And during the test phase, we randomly select an answer from all the replies. Therefore, only one reply should be expected, while the others no. Figure \ref{fig:scenarios} illustrates the test scenario in relation to the training and the original corpus.  

\begin{table}[ht]
\centering
\small
\begin{tabular}{@{}rrrrcrrr@{}}
\toprule 
 \textbf{Metric}  & \textbf{Value}    
\\
\midrule
\textit{Nbr. of utterances)} & 348,442 \\
\textit{Nbr. of utterances per agent} &  49,778 \\
\textit{Nbr. of Dialogues} &  6,138 \\ 
\textit{Avg. nbr. of agents per Dialogue}  & 7   \\
\textit{Avg. nbr. of utterances per Dialogue} &  56 \\
\textit{Avg. length of utterances (words)}  & 13\\
\bottomrule
\end{tabular}
\caption{Multibotwoz Corpus with Dialogue Errors}
\label{tab:corpuserrors}
\end{table}

We designed two set of FSA rules: one for Scenario A and one for Scenario B. While Scenario A does not consider the output of the classifiers, Scenario B does and the set of rules for Scenario A is a subset of rules for Scenario B as described next. Below are the basic rules defined for Scenario A:

\begin{itemize}
\item CR-A1: The user has always  \textbf{permission} to reply to any utterance sent.

\item CR-A2: Whenever an utterance that mentions a participant in the conversation is sent, the mentioned participant has the \textbf{obligation} to reply and the other participants are prohibited.

\item CR-A3: Whenever an utterance is sent from a participant with bot role without any mention, the sender and all other participants with bot role receive a  \textbf{prohibition} to interact.

\item CR-A4: Whenever an utterance is sent from a participant with user role without any mention, participants with bot role that try to interact after a reply is sent to that utterance receive a  \textbf{prohibition}.

\end{itemize}

Scenario B extended Scenario A with the following rules:

\begin{itemize}

\item CR-B1: Whenever an utterance is sent from a participant with user role without any mention and any participant \textit{is expected to reply}, the sender with bot role receives a  \textbf{prohibition} to interact and participants with bot role that try to interact after a reply is sent to that utterance receive a  \textbf{prohibition}.

\item CR-B2: Whenever an utterance that \textit{is expected to be replied} by the participant with user role is sent, the participant with the user role receives an \textbf{obligation} to reply and the other participants are prohibited.

\item CR-B3: Whenever an utterance that \textit{is expected to be replied} by the participant with bot role is sent, the participant with the bot role receives an \textbf{obligation} to reply and the other participants are prohibited.

\end{itemize}

As a result, Scenario A was implemented with the DSL-CR language through the specification of 6 norms and 3 transitions, while Scenario B was implemented by extending Scenario A. The conversation rule CR-B3 required 6 extra norms and 6 extra transitions (one for each service bot). 

Scenario B rules were used in two experiments: B80 and B90. In the former, the threshold for the confidence score of both classifiers in Equation \ref{eq:prediction} was $k_1 = k_2 = 0.8$, while for the later, $k_1 = k_2 = 0.9$. 

\begin{table}[htb]
\centering
\begin{tabular}{@{}rrrrcrrr@{}}
\toprule 
   & \textbf{Accuracy}  
\\
\midrule
\textit{Baseline} & 0.8649 \\
\textit{A-MLE} &  0.8439 \\ 
\textbf{\textit{AC-CNN}}  & \textbf{0.9234}  \\
\textit{Scenario A} &  0.7600 \\
\textbf{\textit{Scenario B80}}  & \textbf{0.9565}\\
\textit{Scenario B90}  & 0.9174\\
\bottomrule
\end{tabular}
\caption{Accuracy.}
\label{tab:accuracy}
\end{table}

Scenario B80 achieved the highest accuracy and the F1 score was $0.9240$. With these results, we can conclude that both AC-CNN model and our model are better than Baseline (\textit{p-value} $<0.01$), however our model is better than AC-CNN (\textit{p-value} $<0.01$).



Through qualitative analysis, we observed that the majority of the errors with our solution was due to the mediation done by \textit{travel\_bot}.
The AC-CNN was not able to learn the interaction of the mediation and we believe that might be because there are less data for these interactions and we could not design a rule which could be used by the FSA because it is hard to determine with a rule when the \textit{travel\_bot} will interact. We would need, for instance, at least two classifiers to help describing the interaction: a multi-topic classifier, i.e., a classifier that can classify more than one topic for a given utterance and a dialog act classifier in order to classify the ones that \textit{travel\_bot} uses to mediate.

\section{Related Work}

End-to-end data-driven dialogue systems have been built and evaluated \cite{Serban18} and some of them were built for multi-party dialogues. However, they were disentangled into dyadic dialogues before the modeling. The most closest work is \cite{Ouchi2016}, in which a model that encoded the context to predict the addressee and a response in multi-party conversation was proposed. However, their approach do not comprise a hybrid approach as ours, in which we model also interaction rules based on dialogue features and context. To the best of our knowledge, we have presented a novel work on this paper which integrates both machine learning with rules to address the turn taking problem. \\\\

\section{Conclusions and Future Work} 

This paper presented a corpus and architecture that integrates turn-taking classifiers based on Maximum Likelihood Expectation (MLE), Convolutional Neural Networks (CNN) and Finite State Automata (FSA). The corpus is a synthect adaptation of the Multi-Domain Wizard-of-Oz dataset (MultiWOZ) to a multiple travel service-based bots scenario with dialogue errors and was created to simulate user's interaction and evaluate the architecture. By simulating the user's interaction from the \textit{multibotwoz} corpus, our experiments show that our solution can improve the performance of the AC-CNN.

As future work, improvements can be done on the expressivity of the DSL-CR language in order to handle more dialogue context information. Furthermore, we plan to include online and reinforcement learning into the architecture, so a chatbot would be able to learn turn-taking during interaction, enabling a self-adaptive behavior on the turn-taking model.

\bibliographystyle{unsrt}
\bibliography{arxiv-sigdial19-bibliography}

\begin{thebibliography}{1}

\bibitem{finch2018}
C.~S. Pinhanez, H.~Candello, M.~C. Pichiliani, M.~Vasconcelos, M.~Guerra,
  M.~Gatti de~Bayser, and Paulo Cavalin.
\newblock Different but equal: Comparing user collaboration with digital
  personal assistants vs. teams of expert agents.
\newblock 2018.
\newblock arXiv:1808.08157.

\bibitem{Emas18}
M.~Gatti de~Bayser, C.~Pinhanez, H.~Candello, M.~A. Vasconcelos, M.~Pichiliani,
  M.~Alberio Guerra, P.~Cavalin, , and R.~Souza.
\newblock Ravel: a mas orchestration platform for human-chatbots conversations.
\newblock In {\em The 6th International Workshop on Engineering Multi-Agent
  Systems (EMAS @ AAMAS 2018)}, Stockholm, Sweden, 2018.

\bibitem{Deepdial18}
M.~Gatti de~Bayser, M.~Alberio Guerra, P.~Cavalin, and C.~Pinhanez.
\newblock Specifying and implementing multi-party conversation rules with
  finite-state-automata.
\newblock In {\em Proc. of the AAAI Workshop On Reasoning and Learning for
  Human-Machine Dialogues 2018}, New Orleans, USA, 2018.

\bibitem{GattideBayser19}
M.~Gatti de~Bayser, P.~Cavalin, C.~Pinhanez, and B.~Zadrozny.
\newblock Learning multi-party turn-taking models from dialogue logs.
\newblock {\em CoRR}, abs/1907.02090, 2019.

\bibitem{Budzianowski18}
Pawe\l{} Budzianowski, Tsung-Hsien Wen, Bo-Hsiang Tseng, Inigo Casanueva,
  Stefan Ultes, Osman Ramadan, and Milica Ga\v{s}i\'{c}.
\newblock Multiwoz - a large-scale multi-domain wizard-of-oz dataset for
  task-oriented dialogue modelling.
\newblock In {\em Proceedings of the 2018 Conference on Empirical Methods in
  Natural Language Processing}, pages 5016--5026, Brussels, Belgium, 2018. ACL.

\bibitem{Sacks1974}
Harvey Sacks, Emanuel~A. Schegloff, and Gail Jefferson.
\newblock A simplest systematics for the organization of turn-taking for
  conversation.
\newblock {\em Language}, 50(4):696--735, 1974.

\bibitem{Harris98}
J.~W. Harris and H~Stocker.
\newblock Maximum likelihood method.
\newblock page 824, 1998.

\bibitem{Serban18}
Iulian~Vlad Serban, Ryan Lowe, Peter Henderson, Laurent Charlin, and Joelle
  Pineau.
\newblock A survey of available corpora for building data-driven dialogue
  systems: The journal version.
\newblock {\em Dialogue \& Discourse}, 9(1), 2018.

\bibitem{Ouchi2016}
Hiroki Ouchi and Yuta Tsuboi.
\newblock Addressee and response selection for multi-party conversation.
\newblock In Kevin~Duh Jian~Su, Xavier~Carreras, editor, {\em EMNLP}, pages
  2133--2143. The ACL, 2016.

\end{thebibliography}

\label{lastpage}
\end{document}